\documentclass{article}
\usepackage{iclr2022_workshop,times}
\usepackage{multirow}

\usepackage{amsmath,amsfonts,bm,amssymb}

\def\eqref#1{equation~\ref{#1}}

\def\1{\bm{1}}

\DeclareMathAlphabet{\mathsfit}{\encodingdefault}{\sfdefault}{m}{sl}
\SetMathAlphabet{\mathsfit}{bold}{\encodingdefault}{\sfdefault}{bx}{n}

\def\gF{{\mathcal{F}}}

\def\gN{{\mathcal{N}}}

\newcommand*{\diff}{\mathrm{d}}

\iclrfinalcopy
\usepackage{url}
\usepackage{graphicx}
\usepackage{float}
\usepackage{booktabs}
\usepackage{subfig}
\usepackage{xspace}
\usepackage{makecell}
\usepackage{pifont}
\usepackage[colorlinks=true, allcolors=blue]{hyperref}
\usepackage{cleveref}
\usepackage[inline]{enumitem}
\newcommand{\yes}{\ding{51}}%
\newcommand{\no}{\ding{55}}%

\newcommand{\eg}{e.g.\@\xspace}
\newcommand{\ie}{i.e.\@\xspace}
\newcommand{\iid}{i.i.d.\@\xspace}
\newcommand{\wrt}{w.r.t.\@\xspace}

\title{Multi-scale Physical Representations for Approximating PDE Solutions with Graph Neural Operators}

\author{Léon Migus\textsuperscript{1,2,3}, Yuan Yin\textsuperscript{1}, Jocelyn Ahmed Mazari\textsuperscript{4}, Patrick Gallinari\textsuperscript{1,5} \\
\textsuperscript{1} Sorbonne Université, CNRS, ISIR, F-75005 Paris, France\\
\textsuperscript{2} INRIA Paris, ANGE Project-Team, 75589 Paris Cedex 12, France\\
\textsuperscript{3} Sorbonne Université, CNRS, Laboratoire Jacques-Louis Lions, 75005 Paris, France\\
\textsuperscript{4} Extrality, 75002 Paris, France\\
\textsuperscript{5} Criteo AI Lab, Paris, France\\
\texttt{leon.migus@sorbonne-universite.fr, yuan.yin@sorbonne-universite.fr}\\\texttt{ahmed@extrality.ai, patrick.gallinari@sorbonne-universite.fr}
}

\begin{document}

\maketitle

\begin{abstract}
Representing physical signals at different scales is among the most challenging problems in engineering. Several multi-scale modeling tools have been developed to describe physical systems governed by \emph{Partial Differential Equations} (PDEs). These tools are at the crossroad of principled physical models and numerical schema. Recently, data-driven models have been introduced to speed-up the approximation of PDE solutions compared to numerical solvers. 
Among these recent data-driven methods, neural integral operators are a class that learn a mapping between function spaces. These functions are discretized on graphs (meshes) which are appropriate for modeling interactions in physical phenomena. In this work, we study three multi-resolution schema with integral kernel operators that can be approximated with \emph{Message Passing Graph Neural Networks} (MPGNNs). To validate our study, we make extensive MPGNNs experiments with well-chosen metrics considering steady and unsteady PDEs.
\end{abstract}

\section{Introduction and motivation}
Principled modeling of physical phenomena gives rigorous and interpretable mathematical models, \eg Differential Equations (DEs). However, solving these equations analytically is impossible in most practical cases. To circumvent that, one could seek for an approximated solution through numerical analysis. When the phenomenon involves information and energy exchange in different ranges, methods with multi-scale modeling, \eg multi-grid (multi-resolution) methods, are proposed for solving DEs. Interactions at different scales are modeled with a pyramidal discretization \citep{bergot:hal-00547319,OMALLEY2018175}
. With the multi-scale modeling, the solvers converge more quickly than single-scale methods \citep{10.2118/182701-PA,passieux:hal-00485747}.

In Deep Learning (DL), many methods have been proposed for approximating PDE solutions on a regular grid at a single scale \citep{um2020sol,thuerey2020deepFlowPred}. 
However, in real world applications, the domain of PDEs is often discretized on meshes (represented by Euclidean graphs where vertices are points in an Euclidean domain and the edges with the distance between those points), where the nodes and the edges represent respectively the physical states and their interaction. In this case, we use Graph Neural Networks (GNNs) instead, \eg \cite{pfaff2021learning,xu2021conditionally}.

Some methods use U-net \citep{Ronneberger2015UNetCN}, \eg \cite{wandel2021learning}, 
to enable long-range interactions on regular grid data. Recently, Multipole Graph Neural Operator (MGNO, \citealp{Li2020MultipoleGN}) introduces a new graph-based method in this category. It learns an operator mapping between two function spaces by a MPGNN with a multilevel graph. However, \cite{Li2020MultipoleGN} only focus on reducing computation cost of long-range correlation, inspired by fast multipole methods.

In this work, we explore new ways to extend the multi-scale modeling capacity with neural networks. We would like to shed light on different numerical schema and understand the reason for the choice of multi-scale schema from the lens of DL. 
We observe in practice that the multi-scale DL models share a similar structure with some of the numerical schemes . For example, both composed of a straight-through downscaling and upscaling process, U-net shares a very similar structure with V-cycle schema in multi-scale numerical analysis \citep{10.1093/gji/ggw352}. We then draw inspiration from discretized multi-resolution schema \citep{10.1093/gji/ggw352} such as W-cycle and F-cycle to propose new architectures. 

In this paper, we propose new multi-scale DL architectures, based on the original MGNO, for learning representation of multi-scale physical signals by approximating the functional spaces of PDEs. They are tested with steady and unsteady physical systems. This opens perspectives to rethink the architecture design for multi-scale problems and help practitioners using U-net to include these numerical schema variants in their study. To the best of our knowledge, this is the first work to explore new architecture designs for multi-scale problems within the Machine Learning (ML) community.

We organize our paper as follows: in section 2, we describe formally Graph Neural Operators and the multi-resolution schema. In section 3, we present our evaluation protocol and ablation study. Finally, in section 4, we conclude our work.

\section{Neural Operator and its graph instantiations}

\paragraph{Neural Operator \citep{Kovachki2021}}aims at learning a map from a function defined over a domain (typically Euclidean space) to another with parameterized models, especially neural networks (NNs). 
The objective is to learn a map $G_\theta:a\in \mathcal{F}\mapsto u\in \mathcal{F}'$, where $\gF, \gF'$ are two function spaces defined on a domain $D$. At each position in the domain $x\in D$, the following iterative transformation is applied:
\[
v_{t+1}(x) = \sigma\Big(W v_t(x) + b(x) +\int_D\kappa_\phi(x, x')v_t(x')\diff\nu(x')\Big)\qquad\forall x\in D
\]
where $\theta = \{W, b, \phi\}$ are parameters. $W, b$ are local affine transformation of $v_t(x)$ at each location $x$. The parameterized kernel $\kappa_\phi$ integrates $v_t(x')$ over all $x'\in D$, with $x'\sim \nu$ and $\nu$ is a (probabilistic) measure over $D$. $\sigma$ is a nonlinear Lipschitz activation function. The iteration starts from $t=0$ with the input function $a = v_0$, the solution is the function at final iteration $T$, \ie $u = v_T$. 

\paragraph{GNO and MGNO \citep{DBLP:journals/corr/abs-2003-03485,Li2020MultipoleGN}.}
Given that integrating over the whole domain $D$ is intractable, one possible simplification is to integrate only in a subdomain around $x$, \ie $s(x)\subset D$. By limiting this subdomain to some \iid sampled neighbors $x'\in\gN(x)\subset s(x)$, we obtain:
\[
\int_D\kappa_\phi(x, x')v_t(x')\diff\nu(x')\approx \int_{s(x)}\kappa_\phi(x, x')v_t(x') \diff\nu(x') \approx \frac{1}{|\mathcal{N}(x)|}\sum_{x'\in \mathcal{N}(x)}\kappa_\phi(x, x')v_t(x')\]
which can be implemented with a message passing graph neural network, called GNO. The graph is  constructed by connecting the point at position $x$ to its neighbors $x'$ such that $\forall x'\in \gN(x), \|x-x'\|{}\leq r$. Here $|\mathcal{N}(x)|$ stands for the total number of the neighbors of $x$. 
However, when modeling long-range interactions is necessary, a large $s(x)$ should be considered in GNO, which comes with an expensive computational cost. 
To overcome this problem, MGNO considers the following multi-scale kernel matrix decomposition of the graph kernel:
\[K \approx K_{1,1} + K_{1,2}K_{2,2}K_{2,1} + K_{1,2}K_{2,3}K_{3,3}K_{3,2}K_{2,1} + \cdots\]
where $K_{l,l'}$ is the kernel from scale $l$ to $l'$, the finest scale is $1$. To implement this multi-scale kernel, several architectural designs are proposed in the following section. 

\begin{figure}[t!]
\vspace{-0.3in}
\centering
    \subfloat[V-MGNO\label{fig:v-cycle}]{\includegraphics[height=1.in]{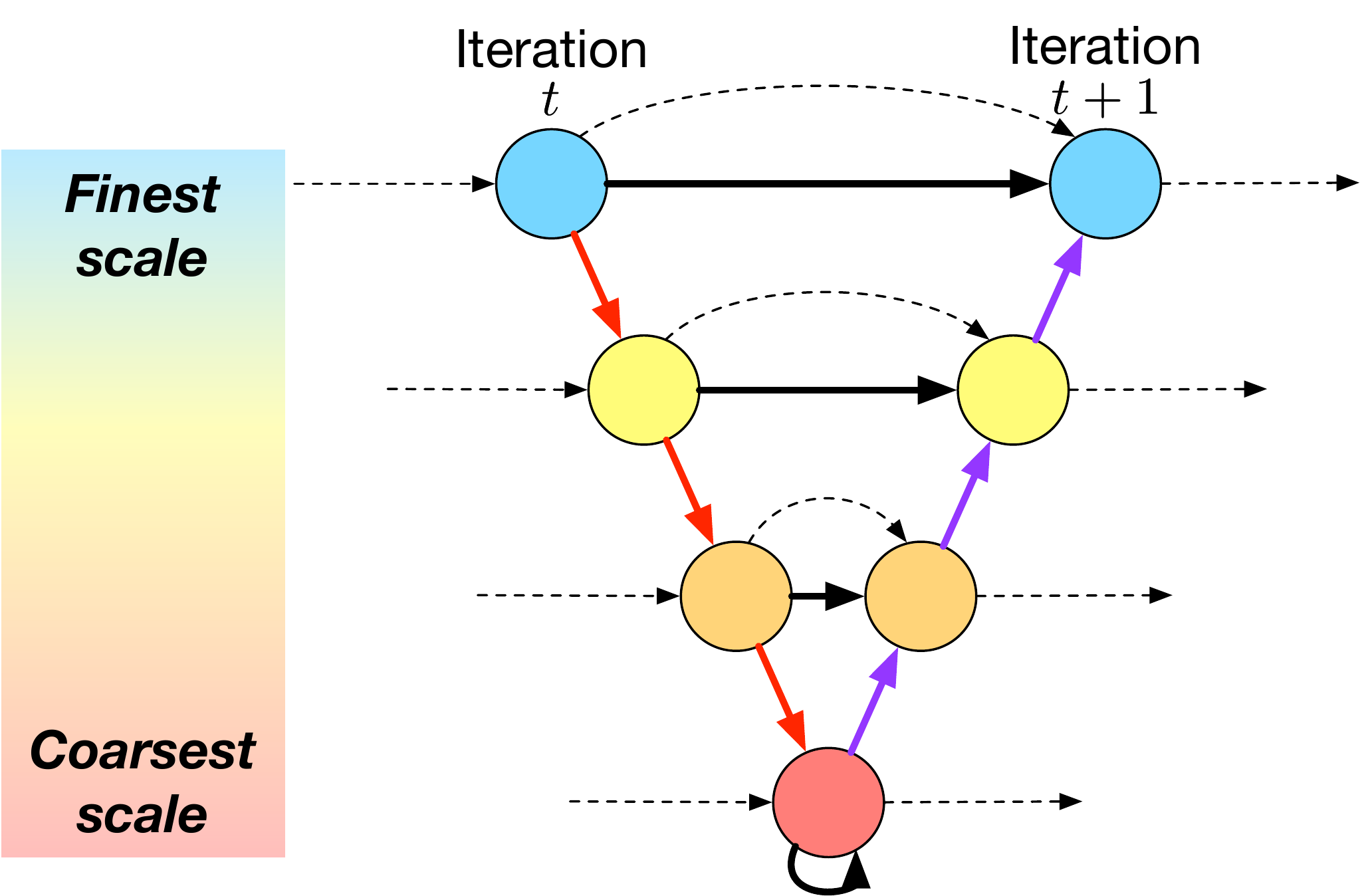}}\hfill
    \subfloat[F-MGNO\label{fig:f-cycle}]{\includegraphics[height=1.in]{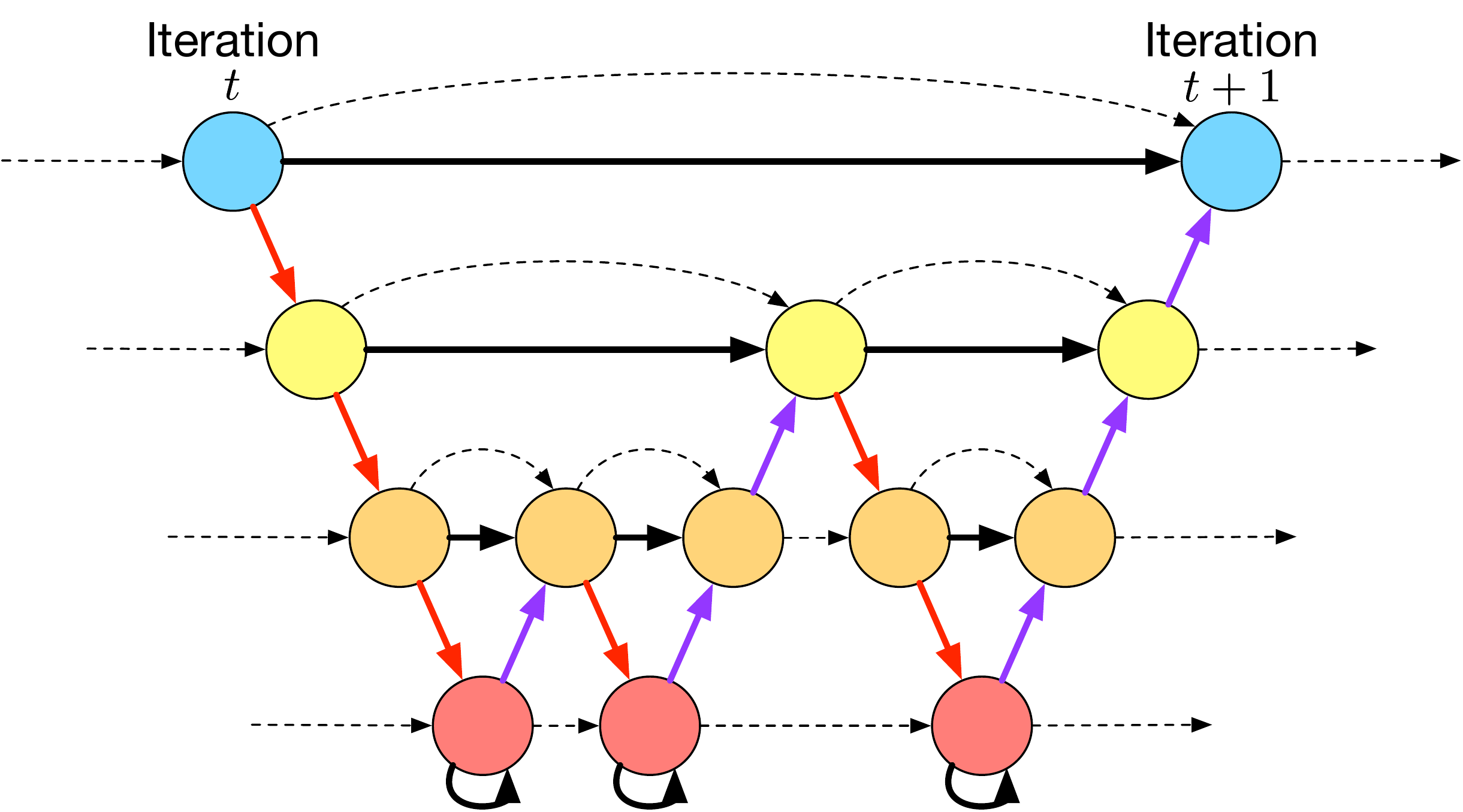}}\hfill
    \subfloat[W-MGNO\label{fig:w-cycle}]{\includegraphics[height=1.in]{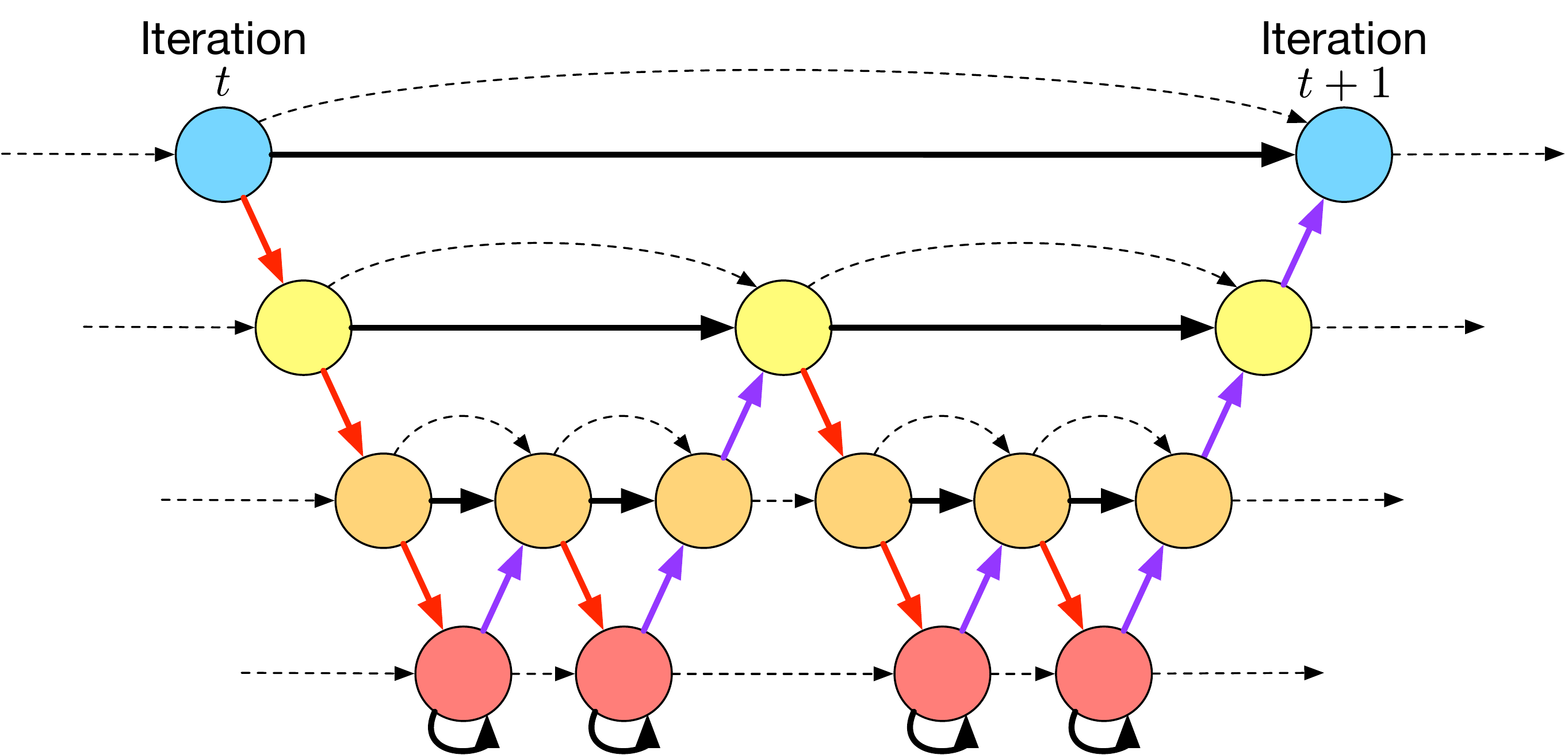}}
    \caption{Illustration of MGNOs with different schema: (a) the original V-MGNO, (b) F-MGNO, and (c) W-MGNO. \textbf{\color{red}\textrightarrow} (red) are downscale kernels, \textbf{\textrightarrow} (black) are in-scale kernels, \textbf{\color{violet}\textrightarrow} (purple) are upscale kernels. $\dashrightarrow$ are skip connections. See Figure \ref{fig:iterative_process} for an example of the iterative process.}
    \label{fig:cycles}
\end{figure}

\paragraph{Multi-scale kernel implementations.}
We present in this section, three architectures for multi-scale kernel implementation: V-MGNO, F-MGNO, and W-MGNO. V-MGNO is proposed in the original paper \citep{Li2020MultipoleGN}. F-MGNO, and W-MGNO are inspired by multi-resolution methods \citep{10.1093/gji/ggw352}. The V-, F-, and W-MGNO are iterative processes and have in common three types of kernels as shown in \Cref{fig:cycles}: downscale, intra-scale, and upscale kernels. The input information is propagated in the multi-scale graph by a cascade of down-scale contraction, intra-scale transformation, and up-scale expansion.

\paragraph{Metrics.}
The goal here is to empirically study the performances of these multi-resolution architectures and their learning stability. We define the following metrics to compare these multi-scale architectures:
\begin{itemize*}
    \item \textit{Number of scales}: indicates the number of scales from the finest to the coarsest one. The choice of the finest and the coarsest scales depends on the discretization schema, and also on the cutoff energy, as well as the computational budget.
    \item \textit{Intra-cycle Kernel Sharing}: indicates whether the kernels are shared inside each iteration. This is because F- and W-MGNO may have several kernels for the same downsampling or upsampling action, \eg between the coarsest two scales, W-MGNO (shown in \Cref{fig:cycles}) need to several downscaling (red arrows) or upscaling (purple arrows). 
    \item \textit{Iteration Kernel Sharing}: indicates if the kernels are shared across all iterations. This allows to study the stability and the complexity of optimizing iterative processes \wrt a given multi-scale architecture.
\end{itemize*}

\section{Experiments}\label{sec:experiments}
We evaluate the performance of our multi-resolution schema F-MGNO and W-MGNO on two families of PDEs, namely 2D steady-state of \textit{Darcy flow} and 1D viscous unsteady \textit{Burgers' equation}.
\paragraph{Datasets.}

\textit{Darcy flow}. We construct our first dataset with the following 2-D steady-state PDE
\[
\begin{cases}
-\nabla \cdot (a(x)\nabla u(x)) = f(x), & x\in (0, 1)^2 \\
u(x) = 0, & x\in \partial (0,1)^2
\end{cases}
\]
$a$ is a random piecewise constant function parameterizing the PDE, given some $f$ a unique $u$ is associated with $a$. In experiments, we approximate the mapping $(a,\nabla a)\mapsto u$.

\textit{Burgers' equation}. In the second dataset, we consider the following 1-D viscous unsteady PDE
\[
\begin{cases}
\partial_t u(x,t) + \partial_x (u^2(x,t)/2) = \nu \partial_{xx} u(x,t), & x \in (0, 2\pi), t \in (0,1] \\
u(x,0) = u_0(x), & x \in (0,2\pi) 
\end{cases}
\]
with periodic boundary conditions. In experiments, we approximate the mapping from the initial condition $u_0$ to the solution at time $t=1$, \ie $u_0 \mapsto u(\cdot,1)$.

For both datasets, 100 examples were generated for train and 100 other ones for validation/test.

\paragraph{Baseline methods.}
We compare our proposed F-MGNO and W-MGNO \wrt GNO and the original V-MGNO. To study to what extent the kernel construction from operator learning standpoint is helpful, proposed architectures are also compared with MLP and GCN \citep{GCN}. 
Methods are categorized into single-scale (MLP, GCN, GNO) and multi-scale (V-, F-, W-MGNO).

\paragraph{On learning stability.} During training, we found out that the multi-scale architectures with Kaiming initialization \citep{DBLP:journals/corr/HeZR015} used in the original paper may lead to divergence in training. This is caused by extremely large loss and gradient at the beginning of the training. One possible explication is that, as the input is transformed through many kernels, an improper initialization will amplify the norm of features along all the forward steps. We therefore chose orthogonal initialization to better control the norm of the output. 

We did a broad hyperparameter search to tune the model. This showed that the initialization was crucial in MGNO architectures, with a main importance for the learning rate and even more for the initialization gain of the orthogonal initialization.

\paragraph{Results.}
We show our results in \Cref{tab:main} for \textit{Darcy flow} and \textit{Burgers' equation}. For all variants of MGNO, we report the results with best architectural metrics.

We found that multi-scales methods outperform single-scale ones in both training and test error. For both datasets, multi-scale methods achieve a decrease of 80-90\% training error compared to the single-scale baselines. This suggests that a better modeling of long-range interaction improves the expressiveness of the neural network. The same improvement tendency was also observed at test time. All variants of MGNO reduce the test error by 10-50\%. This improvement is less significant as it may be limited by number of training data samples.

Among the multi-scale models, for both PDEs, more complex F-/W-MGNO performs slightly better in training, \ie 2.69 with V-MGNO down to 2.23 (-17\%) with W-MGNO for Darcy flow, and 4.25 with V-MGNO down to 3.19 (-25\%) with F-MGNO for Burgers'. However, we did not perceive a significant difference in test error.
We also conducted further ablation studies to compare the impact of architectural metrics. See \Cref{sec:ablation-studies} for details. \\
\paragraph{Discussion.}
To conclude, we confirm with our experiments that multi-scale methods are important for better modeling physical signals, by efficiently maintaining long-range interactions in the spatial domain. Some improvements in training were observed in with F- and W-MGNO, which may indicate an increase in model expressiveness with F- and W-cycle schemes . However, with MGNO implementation, the evidence is not strong enough to support this claim. We suggest further investigation with other graph-based approaches to better understand more complex cycles.  In order to better generalize to unseen data, more constraints could be added to control train/test trade-off.

\begin{table}[t!]
    \caption{Results of our best settings compared to baselines on \textit{Darcy flow} and \textit{Burgers' equation}. We calculate the means and standard deviations for each model based on 4 runs with different seeds. All multi-scale models results are achieved with 4 scales. See \Cref{sec:ablation-studies} for the ablation studies.
    } 
    \centering
    \resizebox{\textwidth}{!}{  
    \begin{tabular}{lllllll}
\toprule
        & &\bf Model & \bf \makecell[l]{Intra-cycle \\ Kernel Sharing} &\bf  \makecell[l]{Iteration \\ Kernel Sharing} &\bf \makecell[l]{Train Error \\ ($\times 10^{-2}$)}&\bf \makecell[l]{Test Error \\ ($\times 10^{-2}$)} \\
        \midrule
      \multirow{6}{*}{\rotatebox[origin=c]{90}{\bf \textit{Darcy flow}}} &\multirow{3}{*}{\rotatebox[origin=c]{15}{\bf Single-scale}}
       &MLP & N/A & N/A & 11.90±0.20 & 12.02±0.40 \\
       & &GCN  & N/A & N/A & 11.87±0.15 & 11.88±0.19\\
          & &GNO & N/A & N/A & 6.45±0.21 & 7.13±0.15 \\
\cmidrule{2-7}
       &\multirow{3}{*}{\bf \rotatebox[origin=c]{15}{\bf Multi-scale}} 
       & V-MGNO & N/A & \no & 2.69±0.18 & \textbf{5.68±0.30} \\
       & &F-MGNO (Ours)  & \no & \no & 2.76±0.15 & 5.80±0.32 \\
       & &W-MGNO (Ours) & \no & \no & \textbf{2.23±0.11} & 5.91±0.28 \\
       \midrule
             \multirow{6}{*}{\rotatebox[origin=c]{90}{\bf \textit{Burgers'}}} &\multirow{3}{*}{\rotatebox[origin=c]{15}{\bf Single-scale}}
       &MLP & N/A & N/A & 41.89±0.40 & 42.07±0.11 \\
       & & GCN  & N/A  & N/A  & 27.88±1.46 & 31.00±1.22 \\
          & &GNO  & N/A  & N/A  & 15.30±0.17  & 53.14±0.86 \\
          \cmidrule{2-7}
       &\multirow{3}{*}{\bf \rotatebox[origin=c]{15}{\bf Multi-scale}} 
       & V-MGNO  & N/A & \yes & 4.25±0.10 & 25.76±0.39 \\
       & & F-MGNO (Ours) & \yes & \yes & \textbf{3.19±0.05} & 25.20±0.20 \\
       & & W-MGNO (Ours) & \no & \yes & 3.47±0.07 & \textbf{24.91±0.37} \\
 \bottomrule
    \end{tabular}
    }
    \label{tab:main}
\end{table}

\section{Conclusion}
In this work, we empirically studied the challenging task of representing physical signals at different scales from DL perspective. To do so, we developed two novel multi-scale architectures inspired by discretized multi-resolution schema \citep{10.1093/gji/ggw352} and a neural integral operator, which is motivated by multipole theory \citep{Li2020MultipoleGN}. To the best of our knowledge, this is the first work that proposes to study and design multi-scale DL architectures from numerical schema standpoint. We proposed two MPGNNs to approximate this neural integral operator and we implemented it via F-cycle and W-cycle schema. The latter are iterative processes and hence are challenging to optimize. We defined a set of metrics to evaluate the learning stability of these iterative processes and their performances. We validated our work on two types of PDEs discretized on graphs.

We argue that this work could open perspectives to study novel multi-scale neural architectures, beyond U-net,  and V-F-W-cycle schema, suitable for multi-scale and/or scarce data. One may consider a further study of discretized multi-resolution schema including the properties of their architectures and optimization procedures. This could help to design more interpretable and efficient architectures. Moreover, we think that studying multi-scale neural architectures from discretized multi-resolution schema's standpoint could help to get insights about the capabilities of multi-scale neural architectures to reproduce some properties of discretized multi-resolution schema.

\section{Reproductibility Statement}
We provide a \href{https://github.com/LeonMigu/multi_scale_graph_neural_operator}{GitHub repository} to reproduce the
experiments. We used NVIDIA 12Go single GPUs to conduct the experiments.

\newpage
\bibliography{iclr2022_workshop}
\bibliographystyle{iclr2022_workshop}

\newpage

\appendix

\section{Pipelines}

Graph nodes are uniformly sampled over the domain with different sample number at each scale. Graph edges in each scale are calculated to all points in the same scale within a given distance. For multi-scale models, edges between scales are also calculated likewise. Input node features are values of the input function $v_0(x)$ and their position $x$. Each model predicts $u(x)$. To avoid divergence in training, we use the orthogonal initialization across all models.

\section{Ablation studies}\label{sec:ablation-studies}
We conducted large-scale ablation studies with results in \Cref{tab:ablation-darcy-flow,tab:ablation-burgers}. We analyze the impact of different metrics on impacts of architectural metrics on prediction errors:
\begin{itemize*}
    \item \textit{The impact of number of scales}: We observe that the more the scales, the lower the training error. This shows an increasing tendency in model expressiveness \wrt scales.
    \item \textit{The impact of kernel sharing}: We observe that sharing parameters may help improving training. However, no major differences in test are noticed when evaluating generalization to test samples.
\end{itemize*}

Note that different variants of MGNO are the same according to number of scales. For 2-scale models, V-, F-, and W-MGNO are equivalent. For 3-scale models, F- and W-MGNO are equivalent.

\begin{table}[h!]
    \caption{Ablation studies for \textit{Darcy flow}.} 
    \centering
    \resizebox{\textwidth}{!}{  
    \begin{tabular}{llllllll}
    \toprule
        Method & Scales &  \makecell[l]{Intra-cycle \\ Kernel Sharing} & \makecell[l]{Iteration \\ Kernel Sharing} & \makecell[l]{Train Error \\ ($\times 10^{-2}$)} & \makecell[l]{Test Error \\ ($\times 10^{-2}$)} & \makecell[l]{Time\\/epoch (s)} & \makecell[l]{N. params \\ (M)} \\
        \midrule
     MLP & 1 & N/A & N/A & 11.90±0.20 & 12.02±0.40 & 2.0 & 0.02 \\

        GCN & 1 & N/A & N/A & 11.87±0.15 & 11.88±0.19 & 34.3 & 0.03 \\

        GNO & 1 & N/A & N/A & 6.45±0.21 & 7.13±0.15 & 13.1 & 0.03 \\
    \midrule

        V/F/W-MGNO & 2 & N/A & \no & 4.76±0.35 & 6.67±0.43 & 25.8 & 11.0\\
        V/F/W-MGNO  & 2 & N/A & \yes & 4.88±0.40 & 6.59±0.24 & 25.9 & 2.74\\
    \midrule

        V-MGNO & 3 & N/A & \no & 3.63±0.21 & 5.77±0.12 & 28.0 & 14.1\\
        V-MGNO & 3 & N/A & \yes & 3.24±0.22 & 6.11±0.32 & 27.8 & 3.55\\
        F/W-MGNO & 3 & \no & \no & 3.03±0.22 & 5.94±0.38 & 33.0 & 19.5 \\
        F/W-MGNO & 3 & \yes & \no & 2.63±0.19 & 6.41±0.24 & 33.5 & 14.1\\
        F/W-MGNO & 3 & \no & \yes & 2.47±0.15 & 5.76±0.25 & 34.5 & 4.90\\
        F/W-MGNO & 3 & \yes & \yes & 2.26±0.13 & 5.87±0.23 & 34.6 & 3.55 \\

    \midrule

        V-MGNO & 4 & N/A & \no & 2.69±0.18 & 5.68±0.30 & 28.3 & 15.8 \\
        V-MGNO & 4 & N/A & \yes & 2.02±0.11 & 5.96±0.22 & 28.3 & 3.95 \\
        F-MGNO & 4 & \no & \no & 2.76±0.15 & 5.80±0.32 & 39.2 & 25.5 \\
        F-MGNO & 4 & \yes & \no & 1.88±0.12 & 5.84±0.18 & 37.1 & 15.8 \\
        F-MGNO & 4 & \no & \yes & 1.78±0.05 & 6.14±0.32 & 36.5 & 6.39\\
        F-MGNO & 4 & \yes & \yes & 1.60±0.13 & 6.18±0.31 & 36.9 & 3.95\\
        W-MGNO & 4 & \no & \no & 2.23±0.11 & 5.91±0.28 & 40.9 & 28.2 \\
        W-MGNO & 4 & \yes & \no & 1.68±0.13 & 6.17±0.28 & 39.2 & 15.8  \\
        W-MGNO & 4 & \no & \yes & 1.85±0.08 & 6.07±0.33 & 39.1 & 7.07\\
        W-MGNO & 4 & \yes & \yes & 1.57±0.10 & 6.29±0.27 & 40.5 & 3.95\\
    \bottomrule
    \end{tabular}
    }
    \label{tab:ablation-darcy-flow}
\end{table}

\begin{table}[h!]
    \caption{Ablation studies for Burgers' equation. } 
    \centering

    \resizebox{\textwidth}{!}{  
    \begin{tabular}{llllllll}
    \toprule
        Method & Scales & \makecell[l]{Intra-cycle \\ Sharing} & Depth sharing & \makecell[l]{Train Error \\ ($\times 10^{-2}$)}& \makecell[l]{Test Error \\ ($\times 10^{-2}$)} & \makecell[l]{Time (s) \\/epoch} & \makecell[l]{N. params\\(M)} \\
        \midrule
     MLP & 1 & N/A & N/A & 41.89±0.40 & 42.07±0.11 & 2.8 & 0.017\\

        GCN & 1 & N/A & N/A & 27.88±1.46 & 31.00±1.22 & 1.9 & 0.025 \\

           GNO & 1 & N/A & N/A & 15.30±0.17  & 53.14±0.86 & 27.1 & 1.10 \\
    \midrule

        V/F/W-MGNO & 2 & N/A & \no & 7.21±0.61 & 25.63±0.53 & 73.1 & 10.91\\
        V/F/W-MGNO & 2 & N/A & \yes & 6.35±0.17 & 25.67±0.46 & 66.5 & 2.74\\
    \midrule

        V-MGNO & 3 & N/A & \no & 4.76±0.24 & 27.22±1.19 & 77.5 & 14.13 \\
        V-MGNO & 3 & N/A & \yes & 4.22±0.14 & 26.65±0.49 & 62.8 & 3.54 \\
        F/W-MGNO & 3 & \no & \yes & 4.59±0.16 & 26.64±1.38 & 77.3 & 4.89 \\
        F/W-MGNO & 3 & \yes & \yes & 4.02±0.15 & 25.29±0.86 & 85.8 & 3.54\\

    \midrule

        V-MGNO & 4 & N/A & \no & 4.67±0.16 & 26.19±0.26  & 70.1 & 15.75 \\
        V-MGNO & 4 & N/A & \yes & 4.25±0.10 & 25.76±0.39 & 74.8 & 3.95 \\

        F-MGNO & 4 & \no & \yes  & 3.59±0.09 & 25.45±0.63 & 89.5 & 6.39\\
        F-MGNO & 4 & \yes & \yes & 3.19±0.05 & 25.20±0.20 & 77.5 & 3.95\\
        W-MGNO & 4 & \no & \yes & 3.47±0.07 & 24.91±0.37 & 100.4 & 7.06\\
        W-MGNO & 4 & \yes & \yes & 3.10±0.03 & 25.61±0.31 & 77.7 & 3.95\\
        
    \bottomrule
    \end{tabular}
    }
    \label{tab:ablation-burgers}
\end{table}

\begin{figure}[h!]
    \centering
    \includegraphics[width =0.99\linewidth]{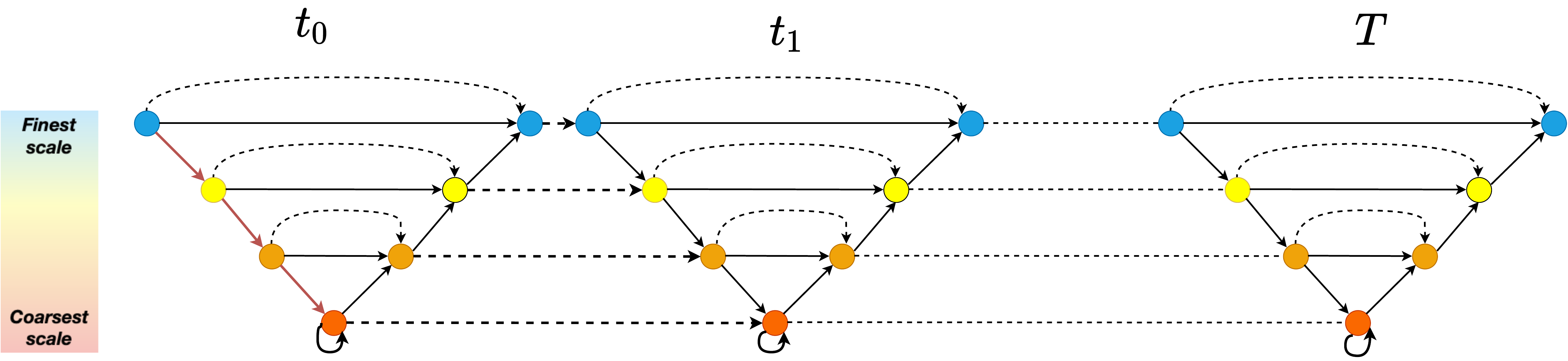}
    \caption{Example of an iterative process with V-cycle multi-resolution architecture. Similar process is applied for F-cycle and W-cycle.}
    \label{fig:iterative_process}
\end{figure}

\end{document}